# Artistic Neural Style Transfer Algorithms with Activation Smoothing

Xiangtian Li *, Han Cao

University of California San Diego, San Diego, 92093, United States, xil160@ucsd.edu

University of California San Diego, San Diego, 92093, United States, h2cao@ucsd.edu

Zhaoyang Zhang

University of California San Diego, San Diego, 92093, United States, zhz088@ucsd.edu

Jiacheng Hu

Tulane University, New Orleans, 70118, United States, jhu10@tulane.edu

Yuhui Jin

California Institute of Technology, Pasadena, 91106, United States, yuhuijin1995@gmail.com

Zihao Zhao

Stevens Institute of Technology, Bellevue, 98004, United States, zzhao7189@gmail.com

**Abstract**

The works of Gatys et al. demonstrated the capability of Convolutional Neural Networks (CNNs) in creating artistic style images. This process of transferring content images in different styles is called Neural Style Transfer (NST). In this paper, we re-implement image-based NST, fast NST, and arbitrary NST. We also explore to utilize ResNet with activation smoothing in NST. Extensive experimental results demonstrate that smoothing transformation can greatly improve the quality of stylization results.

## 1 INTRODUCTION

This paper explores three different Neural Style Transfer (NST) methods. The first NST we explore, proposed by Gatys et al.[1], [2], is called image-based NST. Since their pioneer work, there has been a wealth of research on improving NST performance. Johnson et al.[3]proposed Fast NST, which pre-trains a feed-forward style-specific network and produces stylized results with a single forward pass at the testing stage. Huang and Belongie [4]proposed Adaptive NST to modify conditional instance normalization [5]to adaptive instance normalization. Wang et al.[6]proposed Stylization with Activation smoothinG (SWAG) to apply ResNet in NST. Based on their observation, we conduct extensive experiments on different smoothing transformations, and demonstrate they can all well improve the stylization quality and robustness.

## 2 IMAGE-BASED NEURAL STYLE TRANSFER

The next subsections provide instructions on how to insert figures, tables, and equations in your document.
We start by reimplementing the first NST algorithm proposed by Gatys et al. [1], [2]. Given a content image Ic and a style image Is, the algorithm in [1] seeks a stylized image I that minimizes the following loss:

$$\arg\min_I \mathcal{L}(I_c, I_s, I) = \arg\min_I \alpha\mathcal{L}_c(I_c, I) + \beta\mathcal{L}_s(I_s, I) \quad (1)$$

where $\mathcal{L}_c$ and $\mathcal{L}_s$ represent the content loss and style loss respectively, both are computed based on the layers of VGG-19 [7].

### 2.1 VGG-19

By reconstructing the middle layers of the VGG-19 network, Gatys *et al.* find that a deep convolu- tional neural network is capable of extracting image content and some appearance from well-known artworks.

The paper proposes to get the convolutional filter outputs from different convolutional layers, where we can extract different types of information[8].

### 2.2 Content Loss

The content loss $\mathcal{L}_c$ is defined by the squared Euclidean distance between the feature representations $\mathcal{F}^\ell$ of the content image $I_c$ and the stylized image $I$ in layer $l$, where $I$ is initialized to the content image $I_c$:

$$\mathcal{L}_c(I_c, I) = |\mathcal{F}^\ell(I_c) - \mathcal{F}^\ell(I)|^2 \quad (2)$$

In our experiment, we use the *conv*4_1 layer's filter outputs because *conv*4_2 preserves the content while neglecting the high frequency details.

### 2.3 Style Loss

For style loss $\mathcal{L}_s$, the paper exploits Gram matrix to model the style. Assume that the feature map of a style image Is at layer $l$ in a pre-trained neural network is $\mathcal{F}^\ell(I_s) \in R^{C \times H \times W}$, where $C$ is the number of channels, and $H$ and $W$ represent the height and width of the feature map $\mathcal{F}(I_s)$[9]. Then the Gram matrix $\mathcal{G}(\mathcal{F}^\ell(I_s)') \in R^{C \times C}$ can be computed over the feature map $\mathcal{F}(I_s)' \in R^{C \times (HW)}$ (a reshaped version of $\mathcal{F}(I_s)$):

$$\mathcal{G}(\mathcal{F}^\ell(I_s)') = [\mathcal{F}^\ell(I_s)'][\mathcal{F}^\ell(I_s)']^T \quad (3)$$

The style loss is defined by the square Euclidean distance between the Gram matrix of $I_s$ and $I$:

$$\mathcal{L}_s(I_s, I) = \sum_{l \in \{l_s\}} w_l |\mathcal{G}(\mathcal{F}^\ell(I_s)') - \mathcal{G}(\mathcal{F}^\ell(I)')|^2 \quad (4)$$

where $l_s$ represents some layers in VGG-19, i.e. *conv*1_1, *conv*2_1, *conv*3_1, *conv*4_1, *conv*5_1, and $w_l$ are weights of each layer's contribution to the total loss.

## 3 FAST NEURAL STYLE TRANSFER

Although image-based NST is able to yield impressive stylized images, there are still some limitations. The most concerning problem is the efficiency issue[10]. Usually, image-based NST needs to train for hundreds of iterations to produce a result image, which consumes a lot of time[11].

To address this problem, Johnson et al. [3] propose the first model-based NST algorithm. The idea of the algorithm is to pretrain a feed-forward style-specific network and produce stylized results with a single forward pass at the testing stage. The algorithm of Johnson et al. achieves real-time style transfer. Its architecture is shown in Figure 1 a loss network.

## 3.1 Image Transformation Network

In the training stage, the image transformation network takes in a color image of shape 3 × 256 × 256. Since the image transformation networks are fully convolutional, at the testing stage it can be applied to images of any resolution[12].

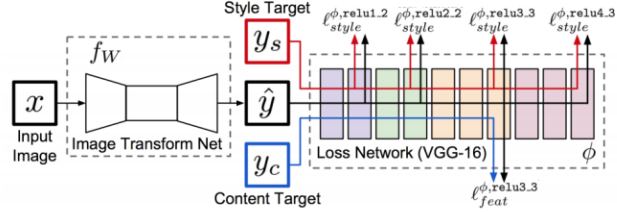

Figure 1. Fast Style Transfer System Overview

## 3.2 Perceptual Loss

We define two perceptual loss functions that make use of a loss network ɸ. In our experiments, the loss network ɸ is the VGG-16 [7] pretrained on ImageNet[13].

**Feature Reconstruction Loss.** The feature reconstruction loss is the squared Euclidean distance between feature representations:

$$l_{feat}^{\phi,j}(\hat{y}, y) = \frac{1}{C_j H_j W_j} |\phi_j(\hat{y}) - \phi_j(y)|_2^2 \quad (5)$$

where y and $\hat{y}$ is the target image and output image respectively, $\phi_j$ is the feature map in $j$-th layers of the VGG-16 network and $C_j \times H_j \times W_j$ is the shape of $\phi_j$.

**Style Reconstruction Loss.** To penalizes the differences in style between $\hat{y}$ and y, we utilize the style reconstruction loss proposed by Gatys *et al.* [2]. Define the Gram matrix $G_j^\phi(x)$ whose elements are given by

$$G_j^\phi(x)_{c,c'} = \frac{1}{C_j H_j W_j} \sum_{h=1}^{H_j} \sum_{w=1}^{W_j} \phi_j(x)_{h,w,c} \phi_j(x)_{h,w,c'} \quad (6)$$

The style reconstruction loss is then the squared Frobenius norm of the difference between the Gram matrices of the output and target images:

$$l_{style}^{\phi,j}(\hat{y}, y) = |G_j^\phi(\hat{y}) - G_j^\phi(y)|_F^2 \quad (7)$$

We define $l_{style}^{\phi,J}$ to be the sum of losses for a set of layers *J* and perform style reconstruction on *J*.

## 4 ARBITRARY STYLE TRANSFER

Based on the previous methods that solve the efficiency issue, we now aim at one-model-for-all, i.e., one single model to transfer arbitrary styles[14]. The algorithm of Huang and Belongie [4] is the first algorithm that achieves a real-time stylization[15].

## 4.1 Adaptive Instance Normalization

Instead of training a parameter prediction network, Huang and Belongie propose to modify conditional instance normalization (CIN) to adaptive instance normalization (AdaIN):

$$\text{AdaIN}\big(\mathcal{F}(I_c),\mathcal{F}(I_s)\big) = \sigma\big(\mathcal{F}(I_s)\big)\left(\frac{\mathcal{F}(I_c)-\mu\big(\mathcal{F}(I_c)\big)}{\sigma\big(\mathcal{F}(I_c)\big)}\right) + \mu\big(\mathcal{F}(I_s)\big) \qquad (8)$$

where $I_c$ and $I_s$ is the content and style image respectively, and $\mathcal{F}$ is the feature representation.

AdaIN transfers the channel-wise mean and variance feature between content and style feature representations.

### 4.2 Network Architecture

Figure 2 shows an overview of the style transfer network based on the AdaIN layer.

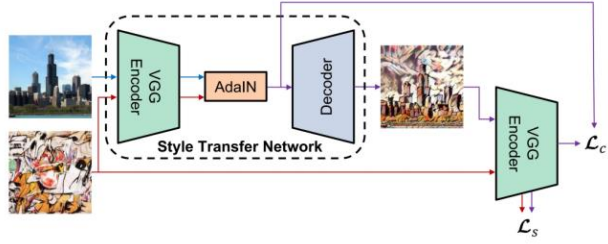

Figure 2. Arbitrary Style Transfer Algorithm Overview

The encoder f is the first few layers of a pretrained VGG-19 [7]. After the encoder, we feed both content and style feature maps to an AdaIN layer that produces the target feature maps t:

$$t = \text{AdaIN}\big(f(I_c), f(I_s)\big) \qquad (9)$$

A randomly initialized decode g, symmetric to the encoder f, is trained to map t back to the image space and get the output image g(t).

### 4.3 Loss Functions

We use pre-trained VGG-19 [7] $f$ to compute the loss to train the encoder. Define the content loss $\mathcal{L}_c$ to be the Euclidean distance between the target features and features of the output image $t$:

$$\mathcal{L}_c = |f\big(g(t)\big) - t|_2 \qquad (10)$$

The style loss matches the mean and standard deviation of the style features:

$$\mathcal{L}_s = \sum_{i=1}^{L} |\mu\big(\phi_i(g(t))\big) - \mu\big(\phi_i(s)\big)|_2 + \sum_{i=1}^{L} |\sigma\big(\phi_i(g(t))\big) - \sigma\big(\phi_i(s)\big)|_2 \qquad (11)$$

where each $\phi_i$ represents a layer in VGG-19. In our experiment we use relu 1_1, relu2_1, relu3_1, relu4_1 layers with equal weights.

By introducing a style loss weight λ we have our final loss:

$$\mathcal{L} = \mathcal{L}_c + \lambda \mathcal{L}_s \qquad (12)$$

## 5 RESNET WITH SMOOTHING TRANSFORMATION

Extensive research in Neural Style Transfer has shown that the correlation between features extracted by a pre-trained VGG network has a remarkable ability to capture the visual style of an image[16]. However, this stylization quality is not robust and often degrades significantly when applied to features from more advanced and lightweight networks, such as the Resnet family[17]. Figure 3 shows an example of style transfer using different models. The VGG transfers style (color, texture, strokes) more faithfully than the ResNet[18].

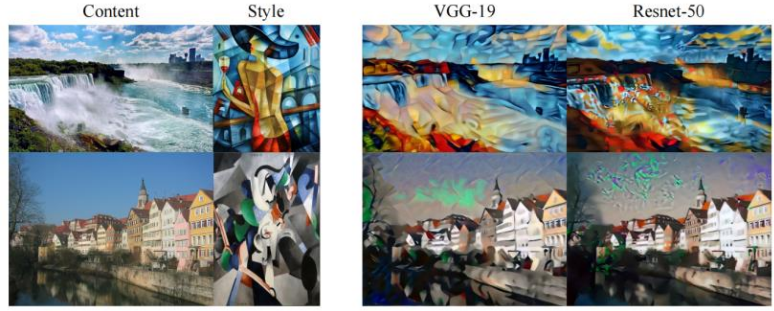

Figure 3. Neural Style Transfer with Different Network Architectures

Wang et al. [6] perform extensive experiments with different network architectures, and find that residual connections, which represent the main architectural difference between VGG and ResNet, produce feature maps of small entropy, which are not suitable for style transfer. They then proposed a novel version of stylization as *Stylization With Activation smoothinG* (SWAG). They propose to avoid peaky activations of low entropy, by smoothing all activations with a softmax-based smoothing transformation. The content loss and style loss are then as follows:

$$\mathcal{L}_{\text{cnet}}(x_0^c, x) = \frac{1}{2} |\sigma\left(F^l(x)\right) - \sigma\left(F^l(x_0^c)\right)|_2^2 \tag{13}$$

$$\mathcal{L}_{\text{sye}}(x_0^s, x) = \sum_l^L \frac{w_l}{4D_l^2 M_l^2} |G^l\left(\sigma\left(F^l(x)\right)\right) - G^l\left(\sigma\left(F^l(x_0^s)\right)\right)|_2^2 \tag{14}$$

where σ is the softmax-based transformation.

In this paper, we extend the smoothing transformation method. Specifically, we find that multiplying a small constant (0.001), the hyperbolic tangent function and the softsign function can reach the similar effect as the softmax function. We set VGG-19 as baseline and conduct experiment on ResNet with and without different types of smoothing transformations. We also conduct experiments to show the limitation of SWAG in stylization and show some failure cases.

## 6 EXPERIMENT

### 6.1 Image-Based Neural Style Transfer

We trained on a Linux (Ubuntu 18.04), with a GeForce 2080 and 12GB memory. Our implementation is based on Python and PyTorch, and Pillow for image processing. We tried two optimizers, Adam and L-BFGS, and L-BFGS produced a better stylized result. However, L-BFGS consumes more memory and takes longer time to get the results[19]. Our actual training was done in 400 iterations for artistic style transfer, with learning rate = 1.

As for the results, we generate artistic style transfer results (Figure 4) with α = 1 and β = 1000, daytime transfer results with α = 1000 and β = 10 and style mixture results with α = 1, β = 1 and the weight of the first layer $w_{conv1\_1} = 5$.

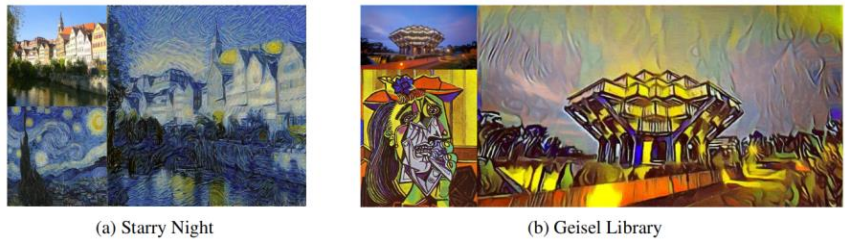

Figure 4. Image-based Neural Style Transfer

## 6.2 Fast Neural Style Transfer

We trained on a Linux (Ubuntu 18.04), with a GeForce 2080 and 12GB memory. Our implementation is based on Python and Pytorch, and Pillow and OpenCV for image and video processing. We train style transfer networks on the MS-COCO dataset [20]. We resize each of the 80k training images to 256 × 256, and train with a batch size of 4 for 40k iterations. We have trained four models using four style images shown in Figure 5a and apply the models on two content images (5b, 14).

We also utilize the model to produce a stylized video (15). What we have done is simply stylizing each frame in the video and combining the frames together. As we stylize each frame independently, the video is not really smooth.

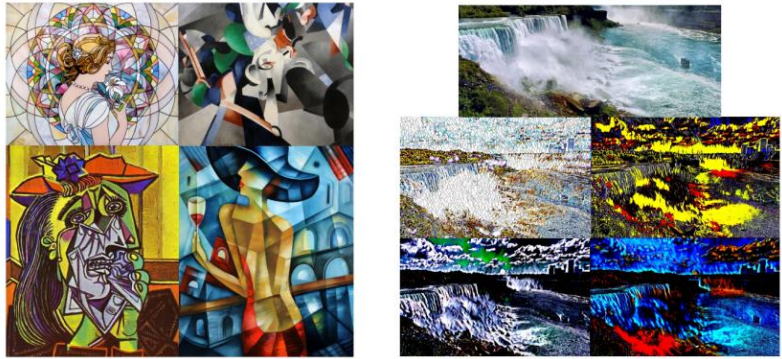

Figure 5. Fast Style Transfer Results

## 6.3 Arbitrary Style Transfer

We trained on a Linux (Ubuntu 18.04), with a GeForce 2080 and 12GB memory. Our implementation is based on Python and PyTorch, and Pillow for image processing. We train style transfer models using MS-COCO dataset [20] as content images and a dataset of paintings collected from WikiArt [9]. We resize each of the 80k training and random crop down images and train with a batch size 8 for 20 epochs.

**Content-style trade-off**. In addition to adjust the weight λ, we can also interpolate at testing stage between feature maps that are fed to the decoder:

$$T(c, s, \alpha) = g\left((1-\alpha)f(c) + \alpha\, AdaIN(f(c), f(s))\right) \tag{15}$$

When α = 0 the model tries to reconstruct the content image and when α = 1 it stylizes the images most. As shown in Figure 6 changing α from 0 to 1.

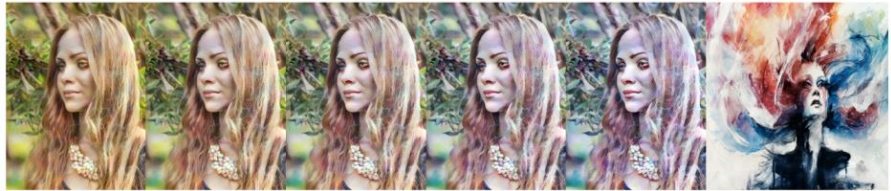

Figure 6. Adaptive Neural Style Transfer with α = 0, 0.25, 0.5, 0.75, 1.0

### 6.4 ResNet with Smoothing Transformation

In this section we discuss an experimental evaluation of the stylization gains of SWAG models. Besides, we find ResNet with SWAG outperforms VGG-19 in some special cases. We also highlight the limitations of SWAG by presenting some failure cases.

We start by evaluating SWAG for ResNet architecture with various smoothing transformations: multiplying a small constant (scale), the softmax function, the hyperbolic tangent function and the softsign function.Specifically, we set the scaling coefficient as 0.001, the weight of the content loss 13 as 1 and the weight of the style loss 14 as $10^{12}$ . The smoothing transformations are applied to layers $conv3_*$ and $conv4_*$.

Figure 7 presents style transfer results on four different images, using the pre-trained versions of all networks, comparing results of SWAG and standard stylization. The results of the standard VGG model are also shown for comparison.

As we can learn from Figure 9, ResNet without SWAG generates results with lowest quality. The different smoothing transformations reach a similar effect. In this case, we would prefer the Softmax, Tanhand Softsign to scaling because they are hyperparameter-free, and achieves similar effects with the scaling smoothing method.

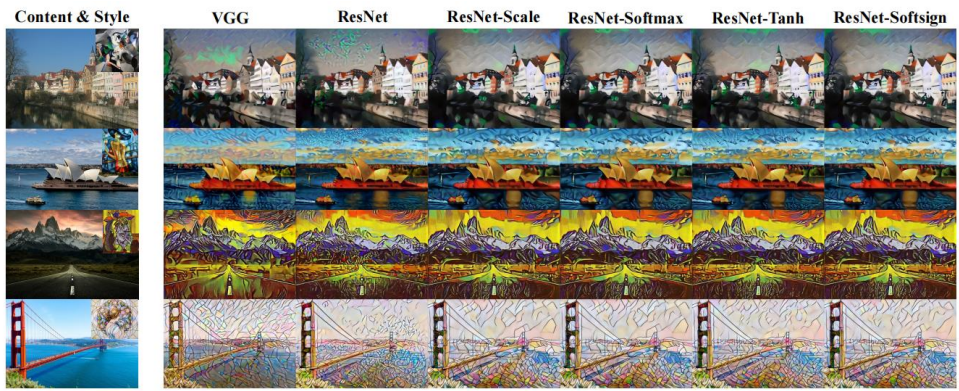

Figure 7. Comparison of Neural Style Transfer Performance Between VGG andResNet with Different Smoothing Transformations. (Please zoom in the picture for a detailed comparison)